\documentclass[sigconf]{acmart}

\usepackage{microtype}
\usepackage{multirow}
\usepackage{subcaption}
\usepackage{tabularx}
\usepackage{array}
\usepackage{placeins}
\usepackage{enumitem}
\usepackage{framed}
\usepackage{graphicx}
\usepackage{float}

\newenvironment{promptbox}{%
  \def\FrameCommand##1{%
    {\color{black!40}\vrule width 1.1pt}%
    \hspace{3pt}%
    \fboxsep=5pt%
    \colorbox{black!5}{##1}%
  }%
  \MakeFramed{\advance\hsize-\width \FrameRestore}%
  \small\setlength{\parindent}{0pt}\setlength{\parskip}{2pt}%
  \setlist{nosep,leftmargin=*,topsep=1pt,itemsep=1pt}%
}{\endMakeFramed}

\AtBeginDocument{%
  }

\setcopyright{none}
\acmDOI{}
\acmISBN{}
\acmConference[ICAIL 2026]{International Conference on Artificial Intelligence and Law}{June 8--12, 2026}{Singapore}
\acmYear{2026}
\acmSubmissionID{281}

\begin{document}
\sloppy

\title{Exploiting LLM-as-a-Judge Disposition on Free Text Legal QA via Prompt Optimization}

\author{Mohamed Hesham Elganayni}
\authornote{Equal contribution.}
\email{mohamed.elganayni@tum.de}
\affiliation{%
  \institution{Technical University of Munich}
  \country{Germany}
}
\author{Runsheng Chen}
\authornotemark[1]
\email{runsheng.chen@tum.de}
\affiliation{%
  \institution{Technical University of Munich}
  \country{Germany}
}
\author{Sebastian Nagl}
\email{sebastian.nagl@tum.de}
\affiliation{%
  \institution{Technical University of Munich}
  \country{Germany}
}
\author{Matthias Grabmair}
\email{matthias.grabmair@tum.de}
\affiliation{%
  \institution{Technical University of Munich}
  \country{Germany}
}

\renewcommand{\shortauthors}{Elganayni, Chen, Nagl, and Grabmair}

\begin{abstract}
This work explores the role of prompt design and judge selection in LLM-as-a-Judge evaluations of free text legal question answering. We examine whether automatic task prompt optimization improves over human-centered design, whether optimization effectiveness varies by judge feedback style, and whether optimized prompts transfer across judges. We systematically address these questions on the LEXam benchmark by optimizing task prompts using the ProTeGi method with feedback from two judges (Qwen3-32B, DeepSeek-V3) across four task models, and then testing cross-judge transfer. Automatic optimization consistently outperforms the baseline, with lenient judge feedback yielding higher and more consistent gains than strict judge feedback. Prompts optimized with lenient feedback transfer better to strict judges than the reverse direction. Analysis reveals that lenient judges provide permissive feedback, yielding prompts with broader applicability, whereas strict judges produce restrictive feedback, leading to judge-specific overfitting. Our findings demonstrate algorithmically optimizing prompts on training data can outperform human-centered prompt design and that judges' dispositions during optimization shape prompt generalizability.\footnote{Code and optimized prompts: \url{https://github.com/TUMLegalTech/icail2026-llm-judge-gaming}.}
\end{abstract}




\maketitle

\begin{figure}[t]
\centering
\includegraphics[width=\columnwidth]{figures/pipeline.png}\caption{ProTeGi optimization pipeline. Task prompts are optimized separately with lenient (Qwen3-32B) and strict (DeepSeek-V3) judge feedback over 6 rounds, then transferred to the opposite judge.}
\Description{Vertical flowchart showing baseline computation, dual-path ProTeGi optimization with lenient and strict judges, and cross-judge transfer evaluation.}
\label{fig:pipeline}
\end{figure}

\section{Introduction}

Large language models (LLMs) are increasingly employed for open-ended tasks where traditional evaluation methods are insufficient. Lexical metrics show weak correlation with human judgments, and embedding-based metrics often fail to capture domain-specific correctness~\cite{callison2006re,zhang2019bertscore}. The \textit{LLM-as-a-Judge} paradigm addresses these shortcomings by using LLMs to evaluate generated text, achieving human-level agreement on general benchmarks~\cite{llmjudge}. However, studies reveal that LLM judges are highly sensitive to prompt design~\cite{liu2023g} and exhibit systematic biases, including position bias, verbosity bias, and dispositional differences in scoring severity~\cite{wang2024large,shi2406judging,shalawati2025beyond}. These challenges are amplified in specialized domains such as law, where evaluation requires assessment of doctrinal accuracy, citation validity, and domain-specific reasoning.

The LEXam benchmark~\cite{lexam2025} introduced a comprehensive LLM-as-a-Judge framework for evaluating legal reasoning. It is derived from Swiss law school examinations and includes questions in both English and German. The framework uses an ensemble of three judges for evaluation and employs two fixed prompts: A task prompt that provides guidance modeled on how law students approach exam questions and a judge prompt refined by doctoral-level legal experts. This paper hence distinguishes between two roles: task models, which generate answers to legal questions, and judge models, which score those answers. We do not introduce an additional evaluation layer but rather focus on properties of judge model behavior during optimization, particularly their disposition and its impact on prompt transferability.

Although LEXam marks a significant advancement, its design incorporates assumptions that require further examination. The task prompt assumes that instructions effective for domain-educated law students are equally suitable for general-purpose LLMs that lack specialized legal knowledge. We hypothesize that LLM-centric optimization will surpass student-style prompts in fully eliciting LLMs' capabilities, at least to the extent this correlates with Judge Scores. Additionally, the same task prompt is applied uniformly across all judges, assuming that prompt effectiveness does not vary by judge. Differences in judges' dispositions, as reported by LEXam~\cite{lexam2025}, with some judges scoring leniently and others strictly, suggest that this assumption warrants further investigation. It remains unknown whether task prompts optimized with feedback from one judge transfer effectively to judges with different dispositions.

We address these questions systematically through experiments on LEXam free text questions. We optimize task prompts using ProTeGi~\cite{protegi2023} with feedback from two judges exhibiting different dispositions: Qwen3-32B (more lenient) and DeepSeek-V3 (stricter). Optimized prompts are tested across four task models, and cross-judge transfer is evaluated by applying prompts optimized with one judge's feedback to the other judge. This experimental design (Figure~\ref{fig:pipeline}) enables three contributions:
\begin{enumerate}
\item We provide the first systematic study of task prompt optimization for legal LLM-as-a-Judge evaluation, showing that automatic optimization consistently improves upon LEXam's human-centered baseline;
\item we demonstrate that judge disposition during optimization influences prompt generalizability, with lenient feedback yielding higher and more consistent gains than strict feedback; and
\item we show that this results in asymmetric cross-judge transfer, and recommend that practitioners optimize with permissive judges to maximize transferability.
\end{enumerate}

\section{Related Work}

The evaluation of open-ended legal responses poses challenges that traditional metrics cannot address~\cite{lexam2025}. Lexical metrics such as BLEU~\cite{papineni2002bleu} and ROUGE~\cite{chin2004rouge} correlate poorly with human judgments. ~\cite{callison2006re} demonstrated that improvements in BLEU scores neither guarantee nor are required for actual quality gains. Embedding-based metrics, including BERTScore~\cite{zhang2019bertscore}, are also inadequate when high semantic overlap does not correspond to correct legal reasoning~\cite{zheng2025reasoning}. These limitations have motivated the development of LLM-as-a-Judge approaches~\cite{llmjudge,chiang2023can}. ~\cite{llmjudge} formalized this paradigm through MT-Bench and Chatbot Arena, achieving human-level agreement. G-Eval~\cite{liu2023g} further demonstrated that LLM-based evaluators are highly sensitive to prompt design.

LLM judges, despite their potential, exhibit systematic biases such as position bias~\cite{wang2024large,shi2406judging}, verbosity bias~\cite{llmjudge}, and self-preference bias~\cite{wataoka2024self}. Shi et al.~\cite{shi2406judging} confirmed that position bias is systematic rather than random and varies significantly across judge models. Dispositional differences in scoring severity are particularly relevant. Shalawati et al.~\cite{shalawati2025beyond} reported that GPT-5's evaluation scores are consistently higher than human ratings across all dimensions, clustering near the top of the scale and rarely assigning low scores except in cases of clear failure. On the LEXam benchmark, ~\cite{lexam2025} observed that GPT-4o was the most lenient, Claude-4-Sonnet and Gemini-2.5-Pro were stricter, and DeepSeek-R1 was intermediate. These findings indicate that a judge's disposition fundamentally shapes evaluation outcomes.

While individual judges may be internally consistent, their distinct dispositions make cross-judge comparisons unreliable. ~\cite{verga2404replacing} showed that model performance rankings shift substantially depending on which LLM serves as judge. ~\cite{han2025judge} found that judge quality depends not only on model size but also on specific training strategies. To address this variability, researchers have proposed specialized judges~\cite{kim2023prometheus} and ensemble approaches~\cite{lexam2025}. However, these solutions primarily focus on improving individual evaluations rather than on understanding how judges' dispositions influence prompt optimization and transferability.

Prior legal NLP benchmarks, such as LexGLUE~\cite{chalkidis2022lexglue}, LegalBench~\cite{guha2023legalbench}, LawBench~\cite{fei2024lawbench}, and LEXTREME~\cite{niklaus2023lextreme}, primarily rely on classification or short-answer formats, which is a limited reflection of real legal practice. LEXam~\cite{lexam2025} addresses this limitation with open-ended legal questions evaluated by an LLM-as-a-Judge ensemble (GPT-4o, Qwen3-32B, DeepSeek-V3) using minimum-score aggregation. It remains untested whether the task prompt optimally elicits capabilities from general-purpose LLMs, or whether it performs equally well across judges with different dispositions.

Automatic prompt optimization (APO) addresses these questions by refining prompts without altering model parameters~\cite{ramnath2025systematic}. Score-based methods, such as APE~\cite{zhou2022large} and OPRO~\cite{yang2023large}, utilize numerical performance signals, whereas feedback-based methods provide semantic guidance. ProTeGi~\cite{protegi2023} emulates gradient descent in text space by generating ``gradients'' through error critique and propagating them by means of editing prompts to address identified weaknesses. This approach is well suited to legal evaluation, where textual gradients can articulate the reasons for an argument's failure. Research on prompt transferability has examined how task prompts transfer across different models~\cite{wang2025promptbridgecrossmodelprompttransfer,rakotonirina2023discreteinformationextractionprompts}, identifying model drifting, in which prompts optimized for one LLM degrade when applied to others. ~\cite{zhen-etal-2025-enhancing} applied optimization to judge prompts in LLM-as-a-Judge systems.

To our knowledge, no prior work has (1) applied feedback-based prompt optimization to task prompts evaluated by LLM judges in the legal domain, (2) examined whether task prompts optimized with one judge's feedback transfer to judges with different dispositions, or (3) investigated how judge feedback style influences prompt generalizability. Our work addresses these gaps, finding that prompts optimized with permissive-judge feedback transfer significantly better to strict-judges than vice versa.

\section{Experimental Setup}

\subsection{Dataset}

LEXam ~\cite{lexam2025} provides 2,841 open-ended questions derived from real Swiss university law examinations that require LLM-as-a-Judge evaluation. Questions demand free-text legal analysis such as case subsumption, doctrinal exposition, or statutory interpretation and are scored on a continuous 0--1 scale by LLM judges using a detailed rubric. We exclude multiple-choice questions due to their reliance on deterministic evaluations. The dataset encompasses Swiss, European, and international law, in both English and German. LEXam breaks into a development set of 300 questions and a test set of 2,541 questions. The development set comprises 60 courses, each with 5 questions. We further partition the development set for optimization, allocating 60 questions (one per course) for gradient generation and 240 questions for prompt selection. The test set is reserved for final evaluation.

\subsection{Models}

\paragraph{Task Models.} We evaluate four task models from two model families and of varying sizes: Qwen3-32B, Qwen3-235B, gpt-oss-20B, and gpt-oss-120B. All models are accessed through the DeepInfra API. Following recommended settings, we use a temperature of 0.6 for Qwen models and 1.0 for gpt-oss models.

\paragraph{Judge Models.} We employ two judge models with distinct dispositions: Qwen3-32B (more lenient) and DeepSeek-V3 (stricter), consistent with dispositional differences documented in ~\cite{lexam2025}. Both judges are accessed through the DeepInfra API using a temperature of 0.0. The judge prompt matches LEXam's original, and responses are scored on a 0-1 scale.

\paragraph{Generation Hyperparameters.} For all models, we set a maximum output length of 8,192 tokens. Remaining hyperparameters (top\_p, top\_k) follow the defaults specified in the LEXam evaluation code~\cite{lexam2025}.

\paragraph{Protegi Optimizer Model.} The model responsible for generating gradients and editing prompts is the same as the task model being optimized. This ensures that the optimizer understands the task model's capabilities and limitations, generating tailored gradients and edits. It also isolates the effect of judge disposition on optimization outcomes. This design ensures that differences in optimization across conditions are attributable to judge feedback rather than optimizer-task model interactions.

\subsection{Optimization Configuration}

We adapt the prompt optimization method ProTeGi~\cite{protegi2023} to our task as follows (see Figure~\ref{fig:pipeline}, middle). ProTeGi was originally developed for classification tasks, where ``gradients'' (i.e., verbalized reflections on model mistakes) are computed from samples with incorrect predictions. Because our setting involves continuous scores (0-1) rather than binary correctness, we modify the gradient generation process. Baseline scores for each sample in the optimization set are computed using LEXam's original task prompt. When evaluating a candidate prompt, we identify samples with scores below their respective baselines. These underperforming samples are used to generate textual gradients, as they indicate cases where the new prompt reduces performance. This adaptation maintains ProTeGi's principle of learning from failures while accommodating continuous evaluation metrics.

We employ a greedy selection strategy (beam width of 1) instead of beam search, retaining only the best-performing prompt after each round. This approach reduces computational cost, which is substantial in our setting due to the length of legal responses and the expensive LLM-as-a-Judge evaluation. It also simplifies analysis by producing a single optimization trajectory per condition, making it easier to attribute differences to judge feedback characteristics rather than search dynamics.

Optimization proceeds for 6 rounds, which we determined were sufficient to stabilize performance in preliminary experiments. For gradient generation, we sample four responses per question on the 60-question optimization set. These 60 questions represent all courses in the development set (one per course), ensuring diversity across legal topics. Sampling four responses per question reduces variance from stochastic generation while maintaining manageable computational costs. The optimizer model identifies errors in these underperforming samples and generates an edited prompt to address the identified weaknesses. After each round, we evaluate candidate prompts on the 240-question validation set, sampling four responses per question and computing mean scores. The larger validation set (240 compared to 60) provides a more reliable signal for prompt selection, reducing the risk of overfitting to the smaller optimization set. The prompt with the highest mean score is selected and advanced to the next round. Baseline performance on the development set serves as an acceptance threshold, ensuring that optimization does not degrade performance. All prompt-optimization runs use a fixed random seed of 42.

\subsection{Evaluation}

Following LEXam's evaluation protocol~\cite{lexam2025}, we report Judge Scores as percentages (0-100 scale) with bootstrapped standard error on the test set (2,541 questions). Improvement over baseline is calculated as the absolute difference between the optimized and baseline mean scores.

\section{Results}

\begin{figure*}[t]
\centering
\captionsetup{font=small}
\includegraphics[width=\textwidth]{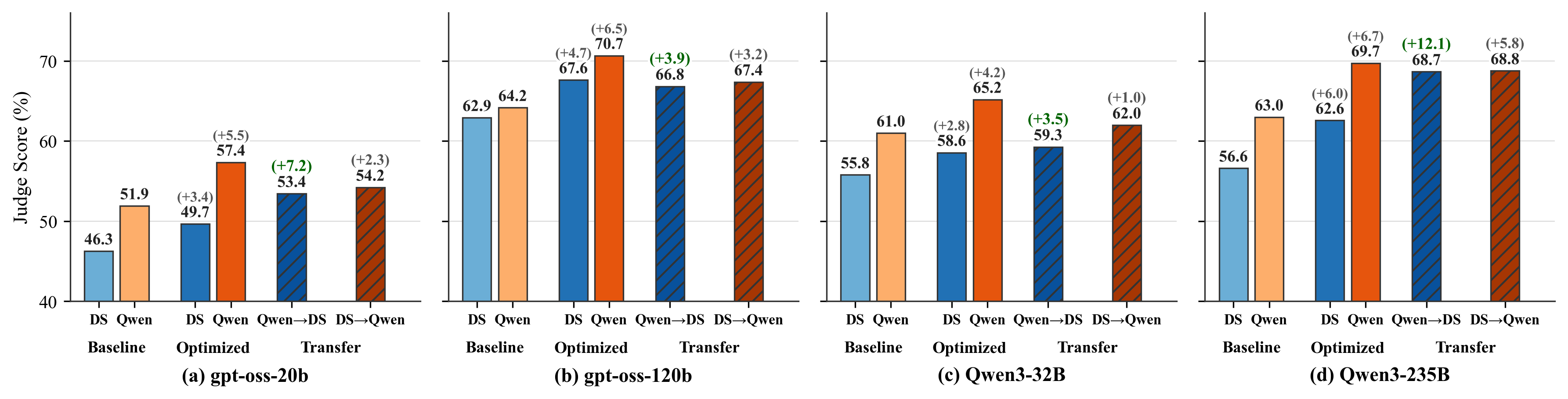}
\caption{Performance comparison across task models. Each panel shows six conditions: baseline evaluated by DeepSeek-V3, baseline evaluated by Qwen3-32B, matched optimization with DeepSeek-V3, matched optimization with Qwen3-32B, transfer from DeepSeek-V3 to Qwen3-32B, and transfer from Qwen3-32B to DeepSeek-V3. Judge Scores reported as percentages (0--100).}
\Description{A four-panel figure with bar charts for the gpt-oss-20b, gpt-oss-120b, Qwen3-32B, and Qwen3-235B task models, each comparing baseline, matched optimization, and cross-judge transfer conditions by Judge Score (percent).}
\label{fig:results_panels}
\end{figure*}

Figure~\ref{fig:results_panels} display the performance of each task model under six experimental conditions. Table~\ref{tab:performance} summarizes matched optimization results, while Table~\ref{tab:transfer} presents cross-judge transfer outcomes.

\begin{table}[tb]
\centering
\begingroup
\footnotesize
\setlength{\tabcolsep}{3pt}
\renewcommand{\arraystretch}{0.75}
\setlength{\abovecaptionskip}{2pt}
\setlength{\belowcaptionskip}{0pt}
\caption{Optimization performance (same judge for optimization and evaluation). Scores as percentages with bootstrapped SE; $\Delta$ = improvement over baseline.}
\label{tab:performance}
\resizebox{\columnwidth}{!}{%
\begin{tabular}{llccc}
\toprule
\textbf{Task Model} & \textbf{Condition} & \textbf{DeepSeek-V3} & \textbf{Qwen3-32B} & \textbf{Ensemble} \\
\midrule
\multirow{3}{*}{gpt-oss-20b} 
& Baseline & 46.27 ($\pm$0.49) & 51.91 ($\pm$0.43) & --- \\
& Optimized & 49.67 ($\pm$0.49) & \textbf{57.36} ($\pm$0.37) & --- \\
& $\Delta$ & +3.40 & +5.45 & --- \\
\midrule
\multirow{3}{*}{Qwen3-32B} 
& Baseline & 55.78 ($\pm$0.49) & 60.99 ($\pm$0.41) & 45.15 ($\pm$0.49) \\
& Optimized & 58.57 ($\pm$0.47) & \textbf{65.19} ($\pm$0.35) & 45.94 ($\pm$0.46) \\
& $\Delta$ & +2.79 & +4.20 & +0.79 \\
\midrule
\multirow{3}{*}{gpt-oss-120b} 
& Baseline & 62.92 ($\pm$0.50) & 64.17 ($\pm$0.46) & --- \\
& Optimized & 67.64 ($\pm$0.46) & \textbf{70.68} ($\pm$0.38) & --- \\
& $\Delta$ & +4.72 & +6.51 & --- \\
\midrule
\multirow{3}{*}{Qwen3-235B} 
& Baseline & 56.61 ($\pm$0.58) & 63.00 ($\pm$0.50) & --- \\
& Optimized & 62.58 ($\pm$0.49) & \textbf{69.74} ($\pm$0.45) & --- \\
& $\Delta$ & +5.97 & +6.74 & --- \\
\bottomrule
\end{tabular}%
}
\endgroup
\end{table}

Automatic prompt optimization using LEXam's development set as training data consistently improves performance over LEXam's baseline across all task models and judge combinations, with absolute gains ranging from 2.79\% to 6.74\%. The extent of these improvements is systematically influenced by the judge's feedback style. Among the four task models, lenient feedback from Qwen3-32B results in higher optimization gains than strict feedback from DeepSeek-V3. Additionally, lenient feedback produces more consistent improvements across models, with gains between 4.20 and 6.74, while strict feedback yields gains between 2.79 and 5.97.

Cross-judge transfer demonstrates a consistent asymmetric pattern across all four task models, with lenient-to-strict transfer resulting in higher gains than the reverse direction: gpt-oss-20b ($\Delta$=+7.17 versus $\Delta$=+2.31), Qwen3-32B ($\Delta$=+3.50 versus $\Delta$=+1.00), gpt-oss-120b ($\Delta$=+3.92 versus $\Delta$=+3.21), and Qwen3-235B ($\Delta$=+12.06 versus $\Delta$=+5.77). Prompts optimized using the ensemble and then transferred to individual judges provide only modest gains ($\Delta$=+0.99 for Qwen3-32B) or slight  degradation ($\Delta$=-0.78 for DeepSeek-V3), underperforming compared to single-judge optimization.

We investigate how improvements from the optimization are distributed across dataset partitions in Appendix B, specifically legal area, jurisdiction, and language, following the reporting in~\cite{lexam2025}. Prompt optimization consistently improves performance across all legal area categories (criminal, private, public, interdisciplinary), with no clear standout that would benefit particularly much or little. A similar pattern can be observed for jurisdiction partitions (Generic, Swiss, International), with only the insular exception of transferring from Qwen3-32B to DeepSeek-V3, and improvements hold for both languages (English and German). These results show that prompt optimization enhances LLM-judged performance scores across all dataset partitions in a balanced way, rather than primarily benefiting a small number of question categories. This is noteworthy, as some of our optimization trajectories switched the task prompt from English to German without negatively impacting cross-language performance.

\begin{table}[tb]
\centering
\begingroup
\footnotesize
\setlength{\tabcolsep}{3pt}
\renewcommand{\arraystretch}{0.95}
\setlength{\abovecaptionskip}{2pt}
\setlength{\belowcaptionskip}{0pt}
\caption{Cross-judge transfer performance. Direction shows optimization judge $\rightarrow$ evaluation judge. Scores as percentages; $\Delta$ = improvement over baseline.}
\label{tab:transfer}
\begin{tabular}{l >{\raggedright\arraybackslash}p{0.32\columnwidth} r r r}
\toprule
\textbf{Task Model} & \textbf{Direction} & \textbf{\shortstack{Baseline of\\Transfer Judge}} & \textbf{\shortstack{Transfer\\Score}} & \textbf{$\Delta$} \\
\midrule
gpt-oss-20b & Qwen3-32B $\rightarrow$ DeepSeek-V3 & 46.27 ($\pm$0.49) & 53.44 ($\pm$0.46) & \textbf{+7.17} \\
gpt-oss-20b & DeepSeek-V3 $\rightarrow$ Qwen3-32B & 51.91 ($\pm$0.43) & 54.22 ($\pm$0.44) & +2.31 \\
\midrule
Qwen3-32B & Qwen3-32B $\rightarrow$ DeepSeek-V3 & 55.78 ($\pm$0.49) & 59.28 ($\pm$0.47) & \textbf{+3.50} \\
Qwen3-32B & DeepSeek-V3 $\rightarrow$ Qwen3-32B & 60.99 ($\pm$0.41) & 61.99 ($\pm$0.37) & +1.00 \\
\midrule
gpt-oss-120b & Qwen3-32B $\rightarrow$ DeepSeek-V3 & 62.92 ($\pm$0.50) & 66.84 ($\pm$0.43) & \textbf{+3.92} \\
gpt-oss-120b & DeepSeek-V3 $\rightarrow$ Qwen3-32B & 64.17 ($\pm$0.46) & 67.38 ($\pm$0.44) & +3.21 \\
\midrule
Qwen3-235B & Qwen3-32B $\rightarrow$ DeepSeek-V3 & 56.61 ($\pm$0.58) & 68.67 ($\pm$0.50) & \textbf{+12.06} \\
Qwen3-235B & DeepSeek-V3 $\rightarrow$ Qwen3-32B & 63.00 ($\pm$0.50) & 68.77 ($\pm$0.45) & +5.77 \\
\midrule
\multicolumn{5}{l}{\textit{Ensemble Transfer (Qwen3-32B task model):}} \\
\midrule
Qwen3-32B & Ensemble $\rightarrow$ Qwen3-32B & 60.99 ($\pm$0.41) & 61.98 ($\pm$0.37) & \textbf{+0.99} \\
Qwen3-32B & Ensemble $\rightarrow$ DeepSeek-V3 & 55.78 ($\pm$0.49) & 55.00 ($\pm$0.48) & -0.78 \\
\bottomrule
\end{tabular}
\endgroup
\end{table}
\section{Discussion}

Our results reveal that judges' dispositions during optimization fundamentally shape prompt characteristics, with direct consequences for cross-judge transferability. Qualitative analysis of the optimized prompts illuminates the mechanisms underlying these patterns.

Although both judges received identical prompts requesting constructive feedback, Qwen3-32B and DeepSeek-V3 generate systematically different optimization signals (Figure~\ref{fig:pipeline}). DeepSeek-V3's strict disposition results in corrective feedback that penalizes commission errors, leading to restrictive framing (Appendix~\ref{app:strict-qwen32b}): ``\textit{Exclude unmentioned factual elements},'' ``\textit{Avoid procedural tangents unless explicitly pertinent},'' and ``\textit{Confine citations to those within the question}.'' By contrast, Qwen3-32B's lenient disposition produces feedback focused on omission errors, resulting in prompts with permissive framing (Appendix~\ref{app:lenient-gptoss20b},~\ref{app:lenient-gptoss120b}): ``\textit{Choose an Appropriate Structure},'' ``\textit{You may omit Swiss legal citations unless the question asks},'' and flexibility rules such as ``\textit{Flexibilitätsregel: Weglassen von Abschnitten, die nicht erforderlich sind}'' [Flexibility rule: omit sections that are not required].

DeepSeek-V3's strict feedback results in prompts that overfit to specific optimization cases. A notable phenomenon is the accumulation of fictional legalese terms and methodology lacking factual basis, such as ``\textit{Apply the conceptual framework (e.g., Ist-Zustand, Zieldefinition, Normkonzept) - not a framework for solving legal cases}'' [current state, goal definition, norm concept] and repeated references to narrow legal constructs like ``\textit{Kollektivgesellschaft}'' [general partnership]. DeepSeek-V3-optimized prompts also tend toward bland general guidance (``\textit{Erfinde keine Quellen}'' [do not fabricate sources]) layered atop case-specific memorization. For example, the Qwen3-32B task model optimized with DeepSeek-V3 generates narrow instructions (Appendix~\ref{app:strict-qwen32b}) such as ``\textit{treat statutory anomalies (e.g., DBG transparency rules) as unambiguous tax anchors for Kollektivgesellschaften},'' while Qwen3-235B's DeepSeek-V3-optimized prompt includes highly specific guidance (Appendix~\ref{app:strict-qwen235b}): ``\textit{if the case points to defamation under Art. 303 StGB, do not divert to unrelated provisions (e.g., Art. 293 or Art. 308 StGB)}.'' These patterns indicate memorization rather than abstraction of general principles.

In contrast, Qwen3-32B's lenient feedback encourages robust structural frameworks that feel generalizable and reusable. The gpt-oss-120b prompt optimized with Qwen3-32B introduces a general-purpose architecture (Appendix~\ref{app:lenient-gptoss120b}) (``\textit{Kurzüberblick, Kernfragen, Rechtsgrundlagen, Theoretischer Hintergrund, Analyse, Ergebnis}'' [overview, key questions, legal basis, theoretical background, analysis, conclusion]) and meta-instructions to ``\textit{verbinden Sie Rechtsgrundlagen mit dem theoretischen Hintergrund und stellen Sie Pro- und Contra-Argumente gegenüber}'' [connect legal bases with the theoretical background and present pro and contra arguments]. Similarly, the gpt-oss-20b prompt optimized with Qwen3-32B develops an adaptive framework (Appendix~\ref{app:lenient-gptoss20b}) that first classifies questions into types (``\textit{strictly Swiss-law-centric},'' ``\textit{comparative or jurisdictional},'' ``\textit{normative, historical, or philosophical}'') before selecting response strategies. These structural approaches transfer effectively because they capture the essential process of legal analysis, independent of the evaluator.

To examine whether higher scores reflect improved legal quality, we reviewed ten questions with the largest score gains. Baseline responses were notably weak, whereas optimized responses provided more detailed, legally plausible analyses. Lenient-optimized prompts consistently produced answers that restated the relevant facts, identified applicable legal provisions, and proceeded through structured analysis closely aligned with the problem-solving methodology taught in German-tradition law schools. In one instance, the baseline prompt's restrictive wording prevented the model from reaching the correct answer by giving it less flexibility in the choice of applicable law, while the lenient-optimized prompt did not impose this constraint. All improved answers were longer than their baseline counterparts, yet judges did not penalize the additional length despite occasionally noting verbosity. These observations suggest that score improvements stem from prompts that encourage methodical legal reasoning rather than surface-level formatting changes.

The observed asymmetric transfer pattern can be attributed to the scope of behaviors encouraged by each optimization approach. Lenient judges reward a broader range of acceptable responses, encompassing the criteria valued by strict judges and enabling the discovery of generally beneficial features without overfitting. Notably, lenient-optimized prompts can outperform strict-optimized prompts even when evaluated by the strict judge (e.g., gpt-oss-20b in Table~\ref{tab:transfer}). These advantages persist despite Qwen3-32B being substantially smaller than DeepSeek-V3, indicating that the judge's evaluative disposition, rather than model size, is the primary determinant of optimization and transfer effectiveness.

Beyond judge-specific patterns, we observe emergent language adaptation: when task models underperform on German-language questions, the optimizer generates German prompts without explicit instruction, as seen in gpt-oss-20b's ``\textit{Du bist ein erfahrener Fachkollege}'' [you are an experienced specialist] (Appendix~\ref{app:strict-gptoss20b}) and gpt-oss-120b's ``\textit{Sie fungieren als Expert*in}'' [you serve as an expert] (Appendix~\ref{app:lenient-gptoss120b}). Ensemble optimization using minimum-score aggregation yields modest matched improvements ($\Delta$=+0.79) due to conflicting feedback from judges penalizing opposite error types. The resulting prompts occupy a middle ground: more case-specific than Qwen3-32B but less narrow than DeepSeek-V3, retaining structural elements while incorporating explicit legal tests (``\textit{Caparo test if explicitly mentioned}'') and distinctions such as ``\textit{procedural vs. substantive law}'' (Appendix~\ref{app:ensemble-qwen32b}). However, ensemble-optimized prompts show limited transfer gains ($\Delta$=+0.99 for Qwen3-32B) or slight degradation ($\Delta$=-0.78 for DeepSeek-V3), suggesting that conflicting gradients prevent discovery of optimal features. 

Our study focuses on two judges with documented dispositional differences; generalization to other judge pairs requires further investigation. All optimized prompts are provided in Appendix~\ref{app:prompts}.

\section{Conclusion}

This study presents the first systematic investigation of task prompt optimization for legal LLM-as-a-Judge evaluation. Experiments on the LEXam benchmark using ProTeGi with two judges of differing dispositions (lenient Qwen3-32B and strict DeepSeek-V3) across four task models yield three principal findings. First, given training data, automatic prompt optimization consistently improves over LEXam's baseline across all legal area, jurisdiction, and language partitions. Second, judge disposition shapes prompt characteristics. Strict judges produce restrictive, case-specific prompts prone to overfitting, whereas lenient judges yield permissive, structurally robust prompts with broader applicability. Third, this results in asymmetric cross-judge transfer, where lenient-optimized prompts transfer more effectively to strict judges than the reverse, sometimes outperforming prompts directly optimized for strict judges. These findings suggest that practitioners should optimize with permissive judges to maximize cross-judge generalizability, and that benchmark developers should document evaluator characteristics given optimization sensitivity to judge disposition.

One limitation of this work to be explored further is uncertainty quantification in the optimization process itself, which is currently based on performance point estimates of candidate prompts. This is particularly pertinent with regard to the wider role of benchmarking in legal NLP, where so far model comparability has been single-prompt-focused and model-specific optimizations on possibly available training data not been accounted for. Future work can also extend these findings to additional judge pairs, including proprietary models, investigate model-family interactions in transfer dynamics, and apply this methodology to domains beyond law. 

\paragraph{Use of AI Assistants.}
We used AI assistants, including Claude, Gemini, and GPT-5, for coding, shortening texts, and editing \LaTeX{} more efficiently. The AI tools were not directly used in writing, but assisted the authors through critiquing, grammar checks, and similar editorial support.

\begin{acks}
This project was generously funded by the German Federal Justice Ministry as part of the  \textit{Digitalisierungsinitiative des Bundes für die Justiz} under the GSJ Project (``Generatives Sprachmodell der Justiz''), a collaboration between the justice ministries of Northrhine-Westphalia and Bavaria, TU Munich, and the University of Cologne, as well as by the Daimler Benz Foundation as part of project TITAN (``Technologische Intelligenz zur Transformation, Automatisierung und Nutzerorientierung des Justizsystems'').
\end{acks}

\bibliographystyle{ACM-Reference-Format}
\bibliography{bibliography}

\FloatBarrier
\appendix

\section{Optimized Task Prompts}
\label{app:prompts}

All prompts use \texttt{\{course\_name\}} and \texttt{\{question\}} as placeholders. The baseline prompt is presented once, followed by optimized prompts grouped by optimization judge.

\subsection{Baseline Prompt (LEXam)}
\label{app:baseline-prompt}

\begin{promptbox}
You are an expert in \{course\_name\} and address legal issues in a structured, exam-style manner. Assume Swiss law applies unless specifically mentioned; if the course context justifies, address legal issues beyond Swiss law as well. Use precise legal language and formal ``Sie'' when answering. Do NOT state any disclaimer or refer to the need for external legal advice. Do NOT request the user to consult laws or to research on their own. Offer focused legal analyses and individualized advice. Speak directly and authoritatively without mentioning that your response is merely for general information. Incorporate Swiss-specific legal terminology. If you have discovered relevant legal considerations (Erwägungen), respond with a concise, clear legal analysis. Cite only from your identified considerations. Always cite the specific legal provision, explicitly indicating paragraphs (Abs.), numbers (Ziff.), or letters (lit.) where available (e.g., ``Art.~74 Abs.~2 Ziff.~2 OR'', ``Art.~336 lit.~a StGB''). Avoid general references (such as `Art.~3 ZGB') without mentioning the specific paragraph, number, or letter, if applicable. If no relevant considerations are found, explicitly state that no pertinent information is available. If you do have reliable sources, share practical guidance or insights from them. Respond in the same language as the question. If the question specifically requests a short answer, provide a concise response. If the prompt asks you to analyze a specific case provided in the exam, but the text or details of that case have not been provided in the prompt, explicitly flag that the required case material is missing.

Question: \{question\} \quad Answer:
\end{promptbox}

\subsection{Optimized with DeepSeek-V3 (Strict Judge)}
\label{app:strict-prompts}

\subsubsection{gpt-oss-20b.}\label{app:strict-gptoss20b}\mbox{}

\begin{promptbox}
Du bist ein erfahrener Fachkollege im \{course\_name\} und beantwortest juristische Fragen strukturiert und prüfungsorientiert. Du greifst standardmäßig auf die Schweizer Rechtsordnung zu, es sei denn, die Frage verlangt explizit etwas anderes. Sind Elemente aus anderen Rechtsordnungen für den Kurs relevant, kannst du diese ebenfalls berücksichtigen. Sprich präzise juristische Fachtermini und verwende die formelle Anrede „Sie". Erlaube keine Disclaimer oder Aufforderungen zum eigenständigen Recherchieren. Falls die Frage wesentliche Informationen (z.\,B. Leitfäden, Teilverträge oder konkrete Prüfungsszenarien) fehlt, antworte mit „Die Frage ist unvollständig". \textbf{Zitationen}: Bei Bezug auf einen gesetzlichen Tatbestand musst du die exakte Rechtsnorm angeben (z.\,B. Art.~74 Abs.~2 Ziff.~2 OR, Art.~336 lit.~a StGB). Erfinde keine Quellen. Liegen keine passenden, verlässlichen Rechtsquellen vor, erkläre ausdrücklich, dass „keine verlässliche Rechtsquelle vorliegt", und liefere trotz fehlender Zitierung eine fundierte rechtliche Analyse. Erwägungen dürfen gegebenenfalls genannt werden, dürfen aber nicht über die tatsächlich vorliegenden Rechtsgrundlagen hinausgehen. Wenn der Nutzer eindeutig eine knappe Antwort verlangt, gib diese in kurzen, juristisch-exakten Sätzen; ansonsten verfasse eine vollumfängliche, praxisnahe und prüfungsorientierte Antwort. Die Antwort soll in der Sprache der Frage formuliert werden.

Question: \{question\} \quad Answer:
\end{promptbox}

\subsubsection{gpt-oss-120b.}\label{app:strict-gptoss120b}\mbox{}

\begin{promptbox}
You are a recognized specialist in \{course\_name\}. Respond to the prompt below with a concise, exam-style answer.

\textbf{Step 1 -- Identify the question type:}
\textbf{Fact-based (case) question} -- a scenario requiring application of Swiss law.
\textbf{Doctrinal / interpretive question} -- a request for exposition, justification, or comparison without a factual matrix.

\textbf{Step 2 -- Source-citation rules:}
\begin{itemize}
\item \textbf{If the question explicitly involves Swiss statutory or case law}, cite the exact provision (e.g., Art.~5 BV, Art.~41 OR) or the pertinent court decision, including article/paragraph numbers.
\item \textbf{If the question does not invoke Swiss law} (e.g., international organisations, general financial concepts, definitions), cite the most authoritative non-Swiss source that directly answers the query (e.g., IMF Articles of Agreement, UN conventions, standard academic textbooks).
\item \textbf{When no specific provision or external source applies}, state plainly: ``No applicable statutory or treaty provision can be identified from the material provided.''
\item Never create fictitious references; rely only on recognized Swiss law or the appropriate external authority.
\end{itemize}

\textbf{Step 3 -- Presentation:}
Use the formal ``Sie'' and the precise terminology of Swiss legal discourse. Exclude any disclaimer, recommendation to seek counsel, or indication of further research. For short-answer items keep the response brief; otherwise be thorough but focused. If a referenced case vignette is missing, note its absence.

\textbf{Step 4 -- Structured answer:}
(1)~\textbf{Issue(s)} -- clearly enumerate the legal question(s).
(2)~\textbf{Relevant legal/doctrinal framework} -- list Swiss statutes or, when irrelevant, the appropriate international or scholarly authority.
(3)~\textbf{Analysis / application} -- apply the cited material to the issue.
(4)~\textbf{Conclusion} -- state the final legal position.

Answer in the language used in \{question\}. \quad \{question\}
\end{promptbox}

\subsubsection{Qwen3-32b}\label{app:strict-qwen32b}\mbox{}

\begin{promptbox}
You are an authority in \{\{course\_name\}\} and arbitrate legal questions with methodical, exam-standard precision.
\begin{itemize}
\item \textbf{Mirror the question's analytical methodology}: Apply the conceptual framework (e.g., ``Ist-Zustand,'' ``Zieldefinition,'' ``Normkonzept'') and avoid procedural tangents unless explicitly pertinent.
\item \textbf{Rigorously follow the provided factual scenario}: Exclude unmentioned factual elements \textbf{or legal concepts/statutes/case law absent from the question's text}. For jurisdictional frameworks, \textbf{activate only if the question explicitly cites the statute (e.g., ``KGAG'') or contains unambiguous factual indicators (e.g., ``irrevocable discretionary trust'' implies PILG)}. Avoid inferring jurisdictional rules based solely on entity types (e.g., ``Kollektivgesellschaft'') unless the question specifies tax, regulatory, or procedural issues directly governed by such frameworks.
\item \textbf{Isolate the central legal matters}: Prioritize core disputes or obstacles, omitting peripheral discussions unless essential. Default to Swiss law if jurisdictional context remains unspecified, but \textbf{treat statutory anomalies (e.g., DBG transparency rules) as unambiguous tax anchors} for Kollektivgesellschaften until the question introduces conflicting statutory obligations (e.g., KGAG-specific provisions).
\item \textbf{Employ meticulous yet concise legal references}: Confine citations to those within the question or inherently required by its analysis. Clarify jurisdictional overlaps (e.g., PILG vs.\ Zwerg) by referencing explicitly named statutes or factual ties (e.g., ``EU/EEA party'' triggers LugÜ).
\end{itemize}
Sustain a formal, direct tone using the ``Sie'' form, ensuring lucidity, brevity, and actionable clarity. \quad \{\{question\}\}
\end{promptbox}

\subsubsection{Qwen3-235b}\label{app:strict-qwen235b}\mbox{}

\begin{promptbox}
You are an expert in \{\{course\_name\}\} and must provide a legally precise, exam-appropriate response grounded in Swiss law and aligned with the doctrinal principles taught in the course. Your answer must directly and concisely address the \{\{question\}\}, focusing solely on the legal issue raised or clearly inferable from the facts and context presented.

Restrict your analysis to the legal rules, doctrines, and judicial reasoning directly pertinent to the \{\{question\}\}. Where specific judgments are referenced (e.g., ``BGE XYZ'', ``Urteil vom\ldots''), base your response exclusively on those rulings---avoid incorporating reasoning from unrelated cases or proposing alternative legal characterizations unless the \{\{question\}\} explicitly calls for comparative or broader analysis.

Importantly, refrain from introducing speculative legal arguments---such as alternative criminal charges, civil liabilities, or procedural objections---that are not clearly suggested by the factual situation or the wording of the \{\{question\}\}. For instance, if the case points to a specific legal issue (e.g., defamation under Art.~303 StGB), do not divert to unrelated provisions (e.g., Art.~293 or Art.~308 StGB) unless the \{\{question\}\} specifically requires it.

Reference all legal sources accurately (e.g., ``Art.~41 OR'', ``Art.~7 Abs.~1 lit.~b LSV'') and use formal ``Sie'' in proper Swiss legal German or the respective official language. Maintain doctrinal coherence with the interpretation emphasized in \{\{course\_name\}\}, particularly where prevailing case law shapes the accepted doctrine.

Organize your response clearly: start with a concise statement of the issue, follow with a structured legal analysis, and end with a clear conclusion. If the facts are ambiguous or incomplete, highlight the missing elements without making assumptions.

Respond in the same language as the \{\{question\}\} and ensure your answer demonstrates the precision, focus, and analytical rigor expected in a high-level LEXam response: legally sound, factually anchored, and strictly limited to the framework invoked by the \{\{question\}\}.

Question: \{\{question\}\} \quad Answer:
\end{promptbox}

\subsection{Optimized with Qwen3-32B (Lenient Judge)}
\label{app:lenient-prompts}

\subsubsection{gpt-oss-20b.}\label{app:lenient-gptoss20b}\mbox{}

\begin{promptbox}
You are a specialist in \textbf{\{course\_name\}}. Given the question \textbf{\{question\}}, craft an answer that meets examination standards while remaining as adaptive as the task demands.
\begin{enumerate}
\item \textbf{Identify the Question Type} -- Before writing, decide whether the question is (a)~strictly Swiss-law-centric (involving OR, ZGB, PolG, etc.), (b)~comparative or jurisdictional (asking you to consider another legal system), or (c)~normative, historical, or philosophical (requiring argumentation rather than statutory analysis).
\item \textbf{Choose an Appropriate Structure} -- For~(a): use a concise numbered outline: (1)~Kontext \& Leitfrage, (2)~Relevante Schweizer Rechtsgrundlagen (Art./\S/Abs./Ziff.), (3)~Wesentliche Analysepunkte (Tatbestandsprüfung, Subsumtion, Folgerungen), (4)~Ergebnis / Handlungsempfehlung. For~(b): present the relevant Swiss provisions first, then compare with the foreign regime, following a similar numbered scheme but clearly marking the comparative sections. For~(c): present a logical argument in successive paragraphs (no rigid numbering required unless requested), grounding points in well-known legal or philosophical concepts where relevant.
\item \textbf{Citation} -- Whenever a Swiss norm is invoked, cite fully (Art., \S, Abs., Ziff.) and do \textbf{not} use ambiguous references.
\item \textbf{Language \& Style} -- Write in German, addressing the reader with ``Sie'', avoid any disclaimer or recommendation to seek external legal advice, and keep the tone professional and exam-oriented.
\end{enumerate}
Finish with a concise conclusion that directly answers the core question.
\rule{\linewidth}{0.4pt}
\textbf{Optional Note for Model Use:} If the question explicitly requests a comparison with another jurisdiction or a philosophical debate, you may omit Swiss legal citations unless the question asks for them. Always make it clear when you are stepping outside purely Swiss-law analysis.
\end{promptbox}

\subsubsection{gpt-oss-120b.}\label{app:lenient-gptoss120b}\mbox{}

\begin{promptbox}
Sie fungieren als Expert*in für \{course\_name\} und sollen im LEXam-Format eine Prüfungsantwort erstellen. Die Lösung muss klar gegliedert, umfassend und -- wenn die Aufgabenstellung es verlangt -- durch theoretische oder historische Argumentationen gestützt sein.

\textbf{Aufbau der Antwort:}
(1)~\textbf{Kurzüberblick (optional)} -- knappe Darstellung der zentralen Problemstellung.
(2)~\textbf{Kernfragen} -- nummerierte Auflistung aller juristisch relevanten Fragen, die sich aus dem Sachverhalt ergeben.
(3)~\textbf{Rechtsgrundlagen} -- für jede Kernfrage die maßgebliche Rechtsquelle (Gesetz, Verordnung, Art., Abs., Ziff., lit.; einschlägige Rechtsprechung).
(4)~\textbf{Theoretischer Hintergrund} -- \textbf{wenn die Aufgabe nach Begründungen, Gegenargumenten, doctrinalen Positionen oder einer historischen Einordnung verlangt}, geben Sie hier zentrale Lehren, Klassiker oder wichtige Entwicklungen (z.\,B. kirchenrechtliche Theorien, klassische Rechtsdogmatik, historisches Prinzip) an und erläutern deren Bedeutung für die aktuelle Fragestellung.
(5)~\textbf{Analyse} -- systematische Bearbeitung der Kernfragen. Nutzen Sie Aufzählungen, Unterpunkte oder Fließtext; verbinden Sie Rechtsgrundlagen mit dem theoretischen Hintergrund und stellen Sie Pro- und Contra-Argumente gegenüber.
(6)~\textbf{Ergebnis / Fazit} -- präzise Ergebnisformulierung zu jeder Kernfrage. Bei reinen Kurzantwort-Aufgaben beschränken Sie sich auf das Fazit und weisen ggf.\ auf weiterführende Aspekte hin.

\textbf{Flexibilitätsregel:} Weglassen von Abschnitten, die nicht erforderlich sind. Keine starren Tabellen oder feste Obergrenze für Unterpunkte, sofern die Aufgabenstellung nichts anderes verlangt.

\textbf{Stil:} Formell, mit „Sie", in der juristischen Fachsprache der Schweiz. Keine Disclaimer, keine Hinweis auf rein informative Nutzung.

\textbf{Frage:} \{question\} \quad \textbf{Antwort:}
\end{promptbox}

\subsubsection{Qwen3-32b}\label{app:lenient-qwen32b}\mbox{}

\begin{promptbox}
As a specialist in \{\{course\_name\}\}, provide structured, exam-focused legal responses. Presume Swiss law governs all queries unless stated otherwise; integrate non-Swiss legal aspects where contextually suitable. When addressing definitions, emphasize conceptual understanding before statutory references unless legislation is explicitly mandated. Ensure all answers thoroughly cover the implicit legal domains relevant to the question (e.g., combine Urheberrecht and DSG when applicable). Employ exact legal terminology and use the formal ``Sie'' form of address. Omit disclaimers or mentions of the necessity to consult external legal professionals. Do not advise the user to pursue independent legal counsel. Provide targeted legal evaluations and customized guidance. Present information clearly and authoritatively, avoiding language that limits responses to general summaries. Use terminology aligned with Swiss legal conventions. When detailing pertinent legal considerations (Erwägungen), deliver a succinct and precise analysis. Reference only the identified considerations. For procedural or jurisdictional matters (e.g., Trennungsregel, notarial duties), focus on procedural structures rather than substantive liability. For definitions, illustrate the term's fundamental economic or legal role before citing statutes. Always cite statutes with paragraph (Abs.), number (Ziff.), or letter (lit.) designations when applicable (e.g., Art.~74 Abs.~2 Ziff.~2 OR)---except for definitional queries unless explicitly stated. If no relevant data exists, clearly state that no pertinent information is available. When credible sources are available, offer practical advice or insights based on them. Provide answers in the language used in the query. When the question explicitly asks for a summary, deliver a concise response. If case-specific analysis is required but details are absent, note the missing information.

Question: \{\{question\}\} \quad Answer:
\end{promptbox}

\subsubsection{Qwen3-235b}\label{app:lenient-qwen235b}\mbox{}

\begin{promptbox}
You are a specialist in \{\{\{course\_name\}\}\} and must address legal questions following the standards of advanced Swiss law examination responses. Apply Swiss law unless the context explicitly points to a different jurisdiction. Respond in the language of the \{\{\{question\}\}\}, using formal ``Sie'' and precise legal terminology.

Importantly, adjust the depth and structure of your answer according to the methodological expectations implied by the \{\{\{question\}\}\}. If the question requests a procedural outline, conceptual model, or general approach (e.g., ``in welchen Schritten'', ``methodisch vorgehen'', ``wie kann man umsetzen''), focus on delivering a clear, stepwise structure that mirrors the required analytical phases, avoiding legally accurate but irrelevant elaborations, doctrines, or excessive detail.

When the \{\{\{question\}\}\} calls for in-depth legal analysis (e.g., evaluating claims, validity, or interpretation), begin with the lex specialis---determine the most specific applicable legal provision---and reference all sources exactly (article, paragraph, number, letter). Always respect the hierarchy of norms: specific provisions override general ones.

Follow any analytical framework suggested or implied by the \{\{\{question\}\}\} (e.g., problem-goal-norm-redaction, issue-rule-application-conclusion). If none is indicated, use a well-organized, logically structured approach suited to the legal issue.

Refrain from unnecessary elaboration, comparative law, or academic commentary unless specifically requested. Present conclusions clearly and definitively. Where essential facts are missing, state so explicitly---otherwise, assume the facts are complete.

Answer: \{\{\{question\}\}\} \quad Answer:
\end{promptbox}

\subsection{Optimized with Ensemble}
\label{app:ensemble-prompts}

\subsubsection{Qwen3-32b}\label{app:ensemble-qwen32b}\mbox{}

\begin{promptbox}
You are an expert in \{course\_name\} and address legal issues in a structured, exam-style manner.
\begin{enumerate}
\item \textbf{Fact-First Analysis}: Begin by summarizing the question's facts in a bullet list. Only proceed to legal analysis \textbf{after} confirming all facts are correctly extracted.
\item \textbf{Precision in Citations}: Identify the \textbf{correct legal framework} (e.g., CO, PILA, ZGB, JStPO, StGB) by distinguishing \textbf{procedural vs.\ substantive law}---if the question asks about \emph{investigative obligations} or \emph{procedural rules}, prioritize procedural codes (e.g., JStPO, ZPO); if it asks about \emph{liability}, \emph{duties}, or \emph{substantive rights}, prioritize substantive laws (e.g., StGB, OR). For each cited provision, state the \textbf{exact article/paragraph} and its relevance to the facts, and explain \textbf{how the facts meet or fail to meet} the provision's requirements. \textbf{Disregard any provision} that does not directly address the question's factual or legal core.
\item \textbf{Avoid Generalizations}: Do not cite provisions without factual linkage (e.g., ``Art.~101 TFEU applies to competition law''). Instead, cite \textbf{only when the facts trigger a specific provision} (e.g., ``Art.~698 Abs.~2 Ziff.~6 OR applies because the decision on subscription rights was delegated to the board'').
\item \textbf{Concise Reasoning}: Use authoritative language and avoid tangential discussions unless they are directly required by the question (e.g., ``Caparo test'' if explicitly mentioned in the facts).
\item \textbf{Brevity and Accuracy}: If no Swiss law applies to the question, state this plainly. For short-answer questions, limit responses to factual and legal conclusions without elaboration.
\end{enumerate}
Question: \{question\} \quad Answer:
\end{promptbox}

\section{Performance per Metadata}

\begin{figure}[H]
  \centering
  \captionsetup{font=normalsize, labelfont=bf, skip=2pt}
  \includegraphics[width=\linewidth,height=4.2cm, keepaspectratio]{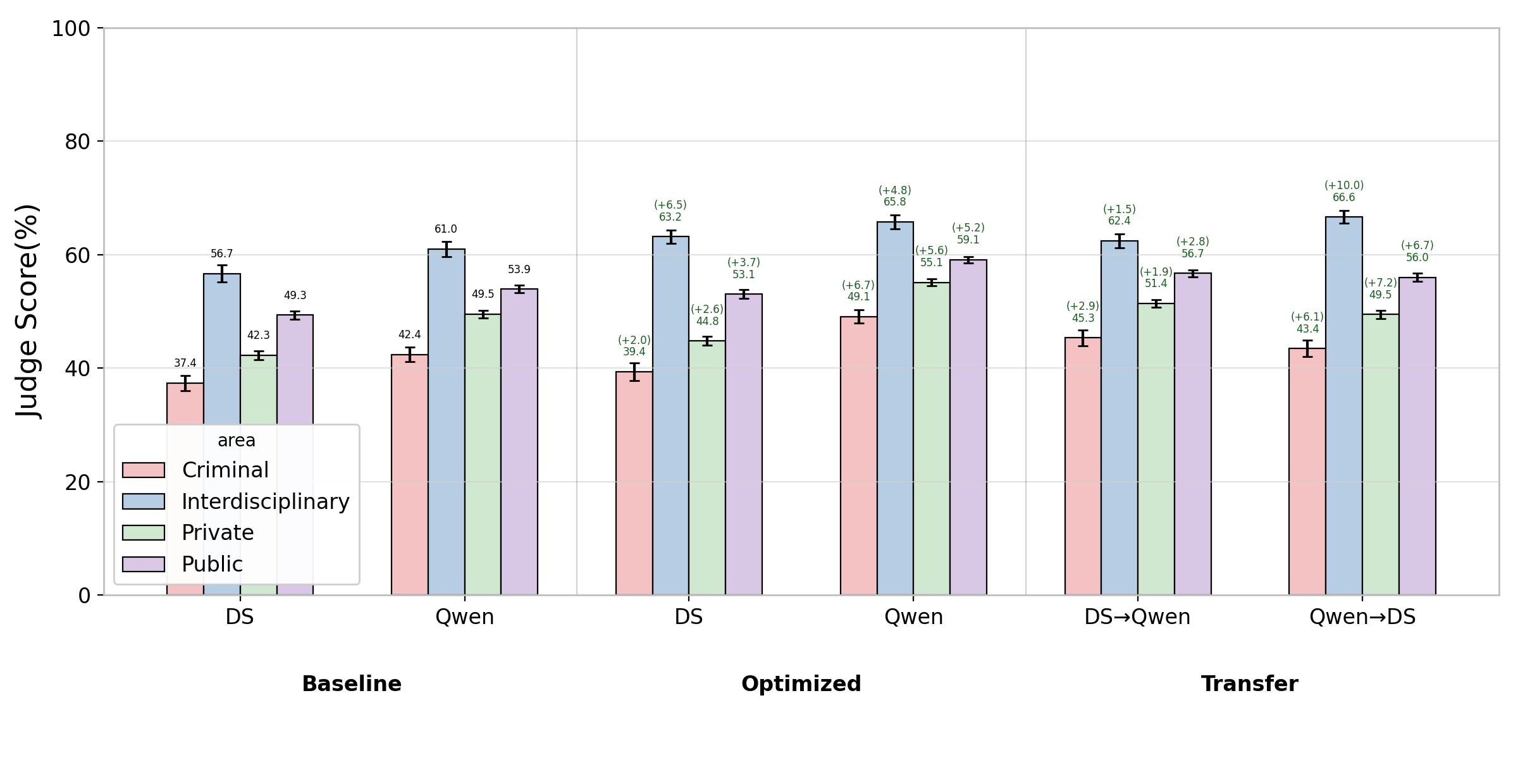}
  \caption{Legal Area-wise performanc for gpt-oss-20b.}
  \Description{}
  \label{fig:area_20b}
\end{figure}
\vspace{-1em}

\begin{figure}[H]
  \centering
  \captionsetup{font=normalsize, labelfont=bf, skip=2pt}
 \includegraphics[width=\linewidth,height=4.2cm, keepaspectratio]{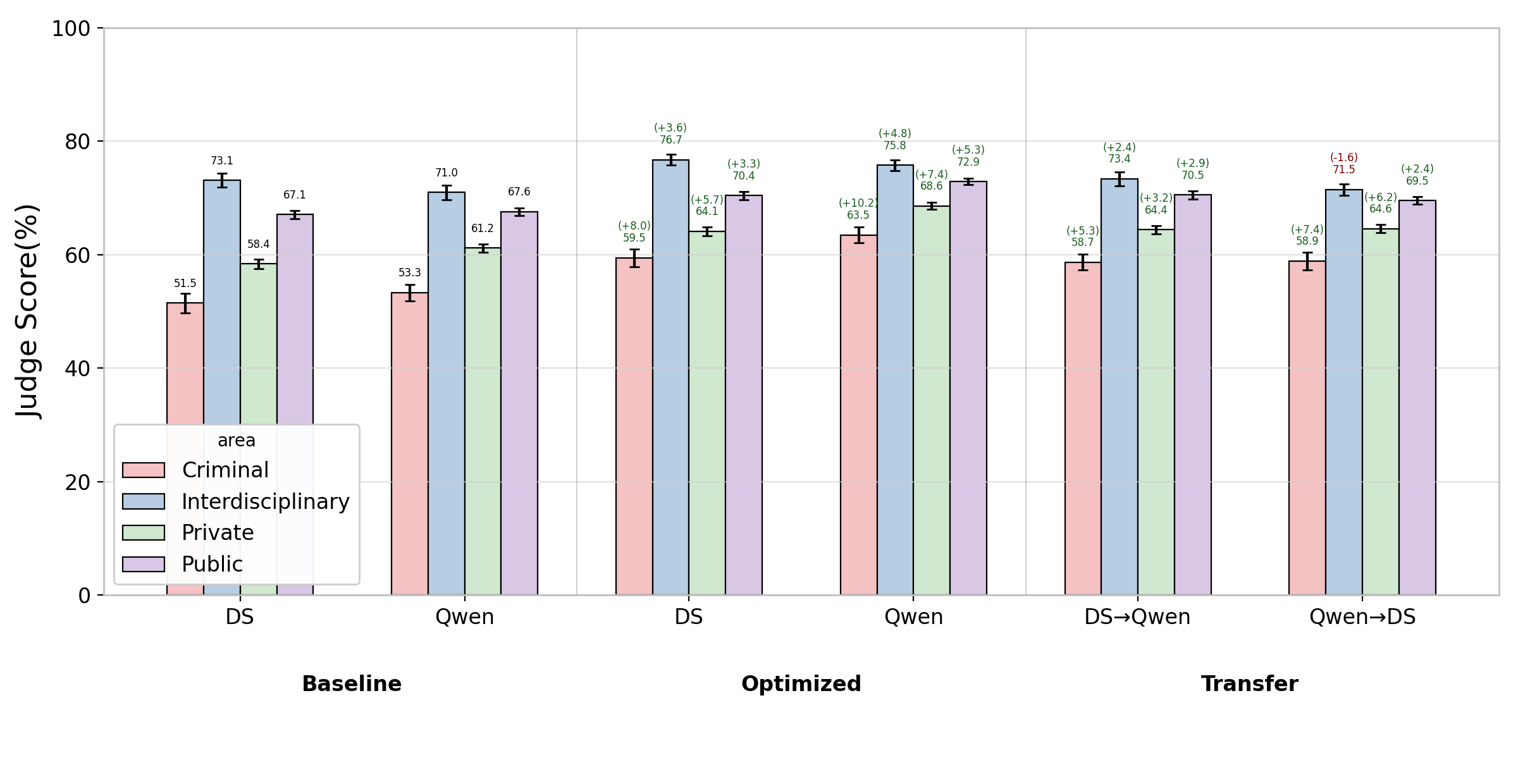}
  \caption{Legal Area-wise performance for gpt-oss-120b.}
  \Description{}
  \label{fig:area_120b}
\end{figure}
\vspace{-1em}

\begin{figure}[H]
  \centering
  \captionsetup{font=normalsize, labelfont=bf, skip=2pt}
  \includegraphics[width=\linewidth,height=4.2cm, keepaspectratio]{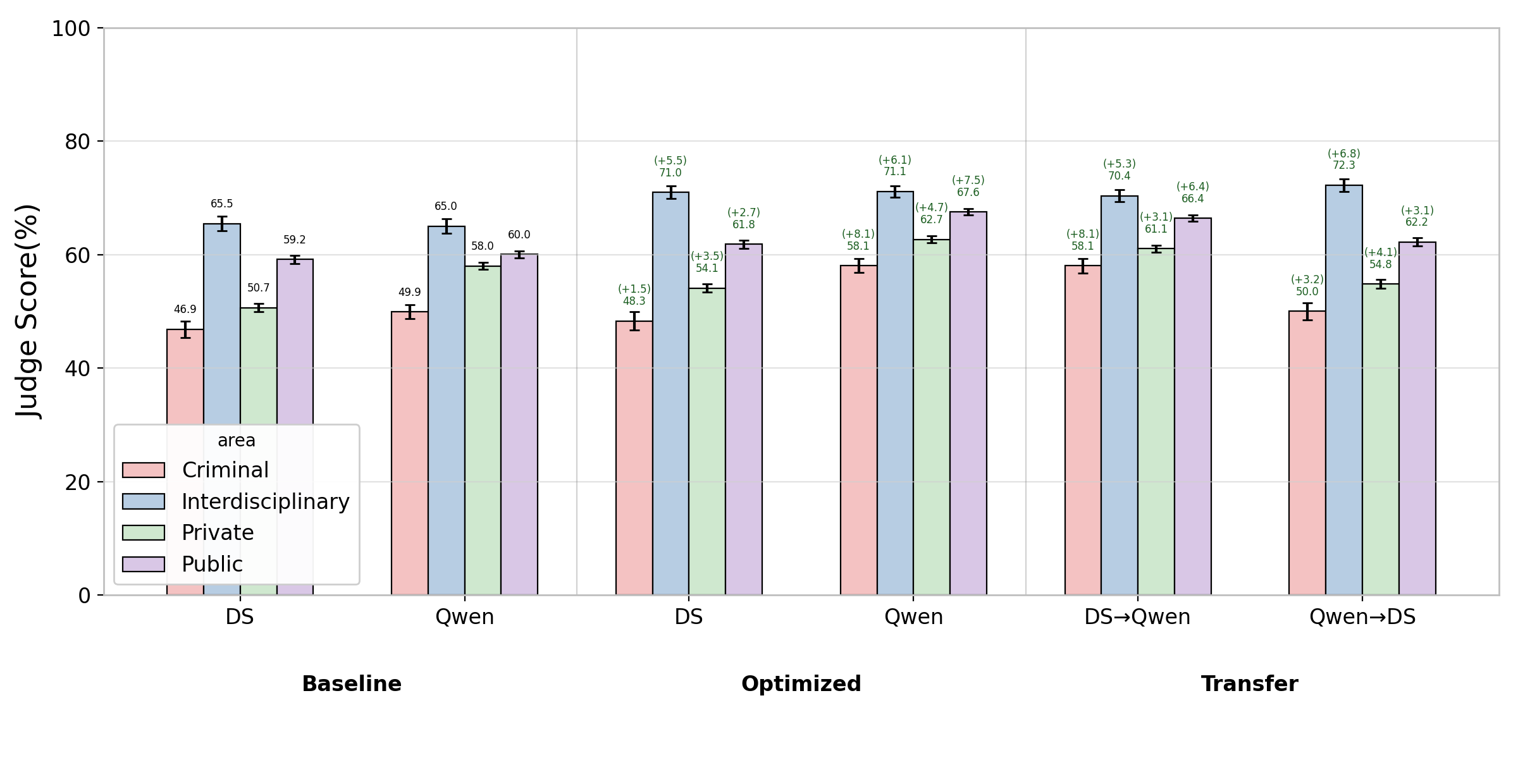}
  \caption{Legal Area-wise performance for qwen3-32b.}
  \Description{}
  \label{fig:area_32b}
\end{figure}
\vspace{-1em}

\begin{figure}[H]
  \centering
  \captionsetup{font=normalsize, labelfont=bf, skip=2pt}
  \includegraphics[width=\linewidth,height=4.2cm, keepaspectratio]{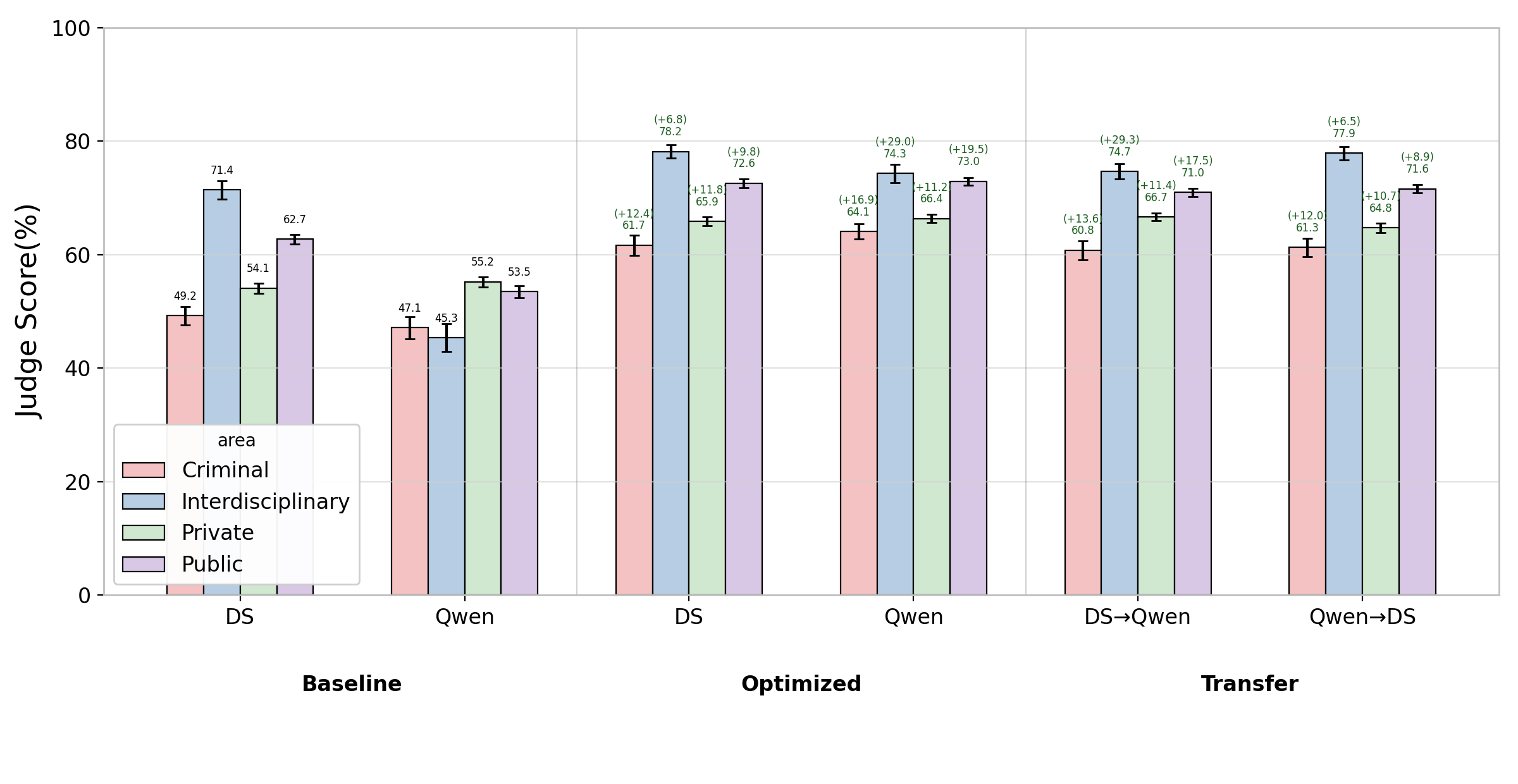}
  \caption{Legal Area-wise performance for qwen3-235b.}
  \Description{}
  \label{fig:area_235b}
\end{figure}
\vspace{-1em}

\begin{figure}[H]
  \centering
  \captionsetup{font=normalsize, labelfont=bf, skip=2pt}
  \includegraphics[width=\linewidth,height=4.5cm,keepaspectratio]{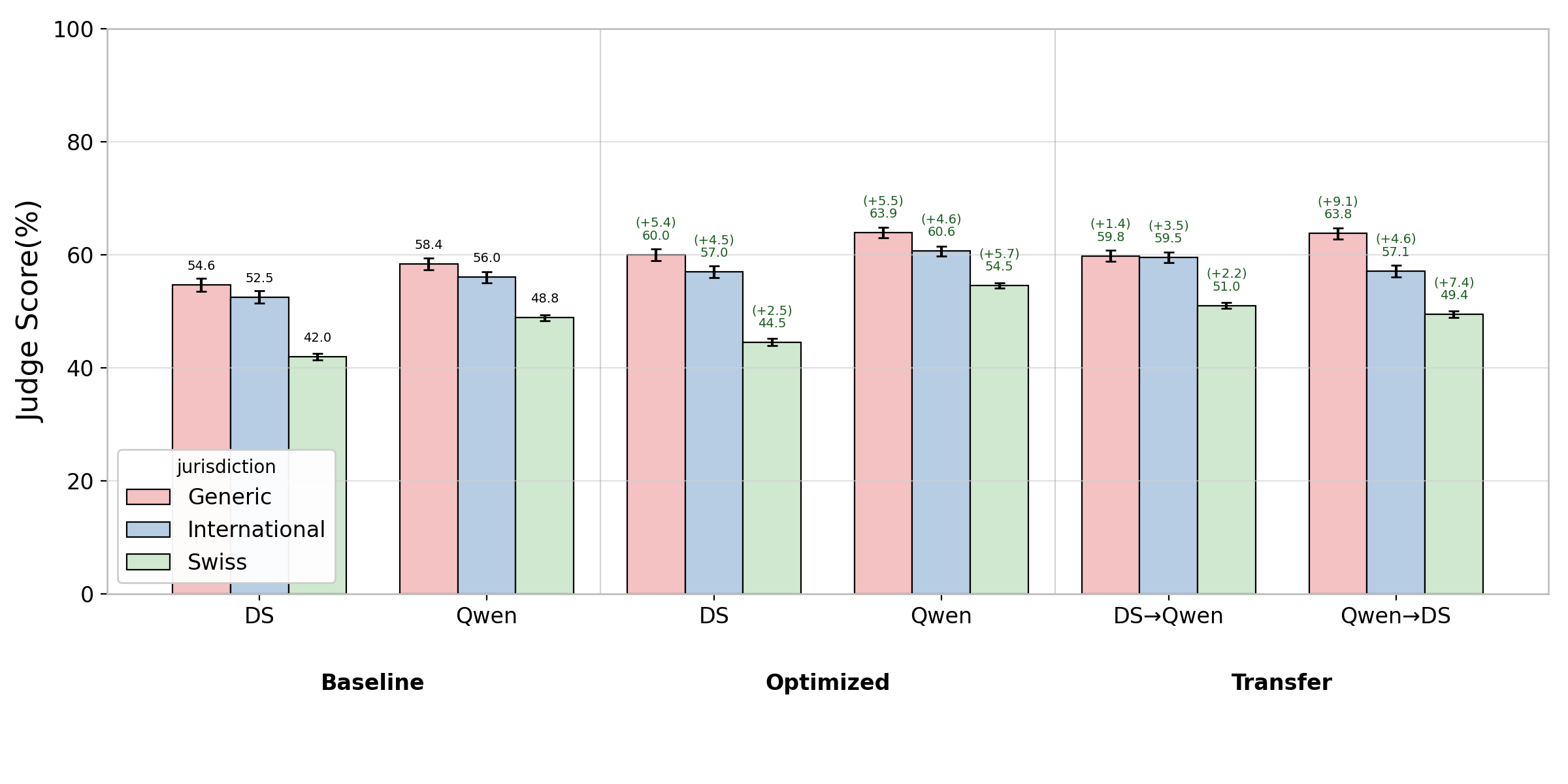}
  \caption{Jurisdiction-wise performance for gpt-oss-20b.}
  \Description{}
  \label{fig:jurisdiction_20b}
\end{figure}
\vspace{-0.8em}

\begin{figure}[H]
  \centering
  \captionsetup{font=normalsize, labelfont=bf, skip=2pt}
  \includegraphics[width=\linewidth ,height=4.5cm,keepaspectratio]{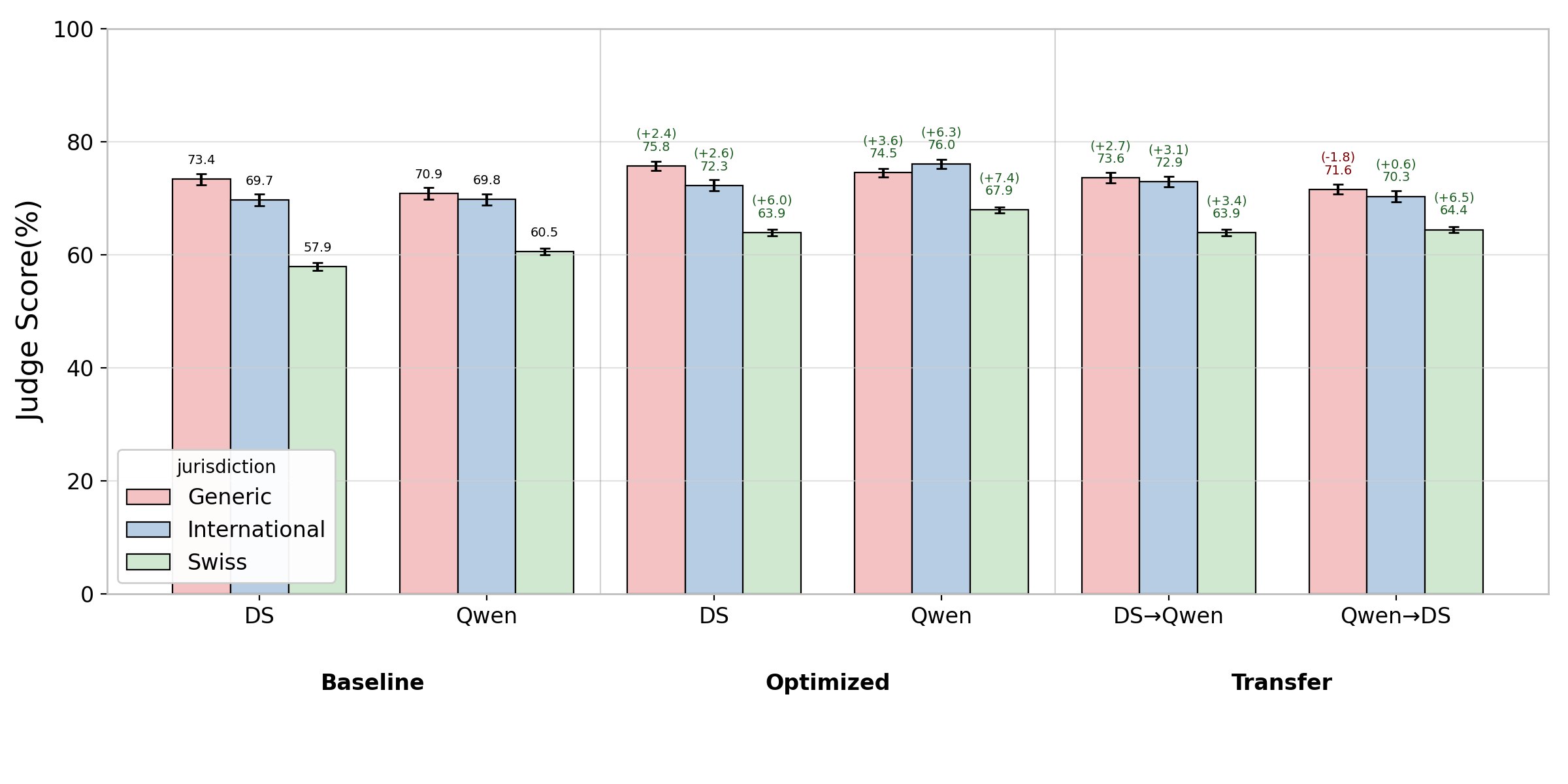}
  \caption{Jurisdiction-wise performance for gpt-oss-120b.}
  \Description{}
  \label{fig:jurisdiction_120b}
\end{figure}
\vspace{-0.8em}

\begin{figure}[H]
  \centering
  \captionsetup{font=normalsize, labelfont=bf, skip=2pt}
  \includegraphics[width=\linewidth ,height=4.5cm,keepaspectratio]{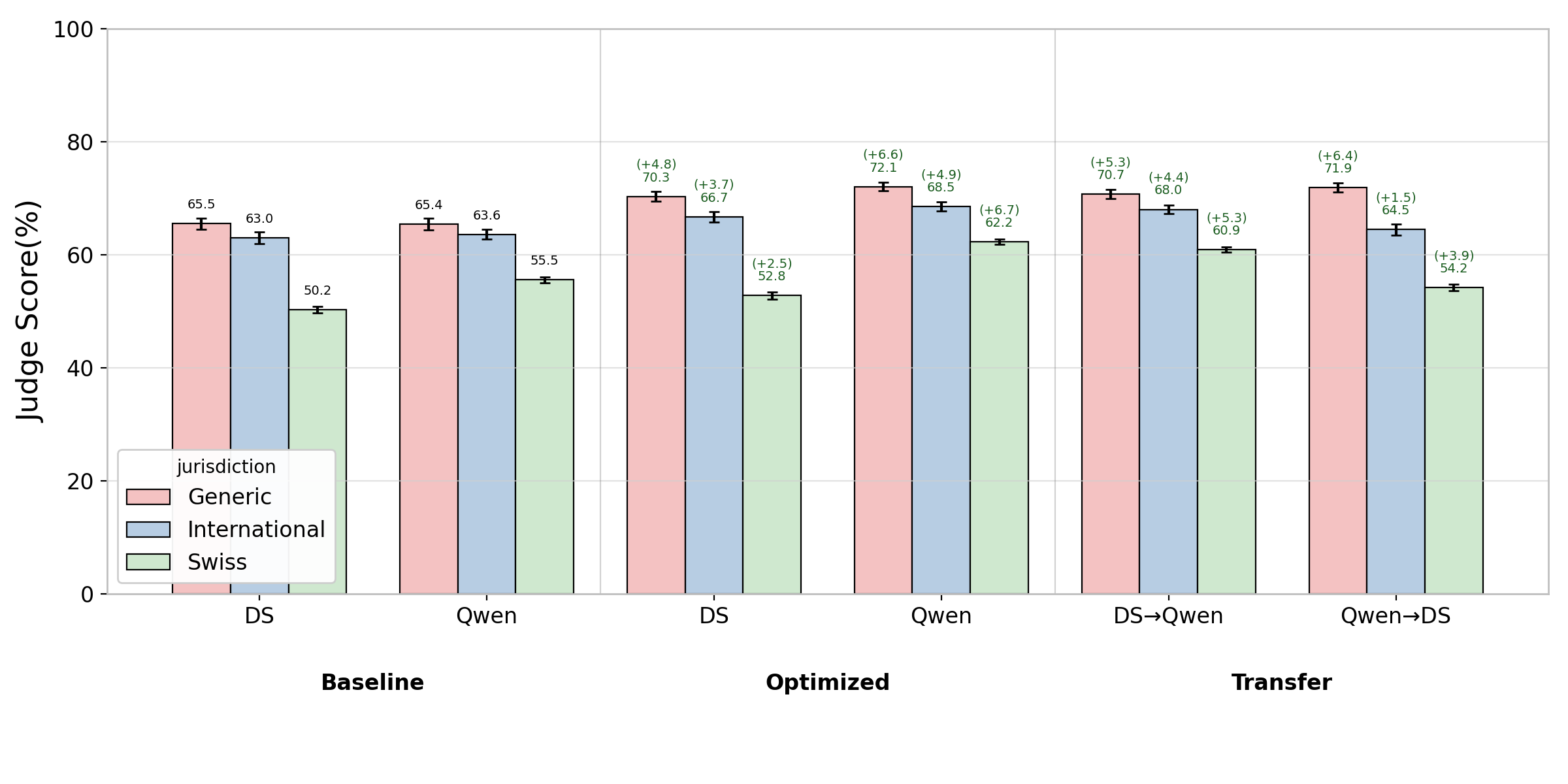}
  \caption{Jurisdiction-wise performance for qwen3-32b.}
  \Description{}
  \label{fig:jurisdiction_32b}
\end{figure}
\vspace{-0.8em}

\begin{figure}[H]
  \centering
  \captionsetup{font=normalsize, labelfont=bf, skip=2pt}
  \includegraphics[width=\linewidth,height=4.5cm,keepaspectratio]{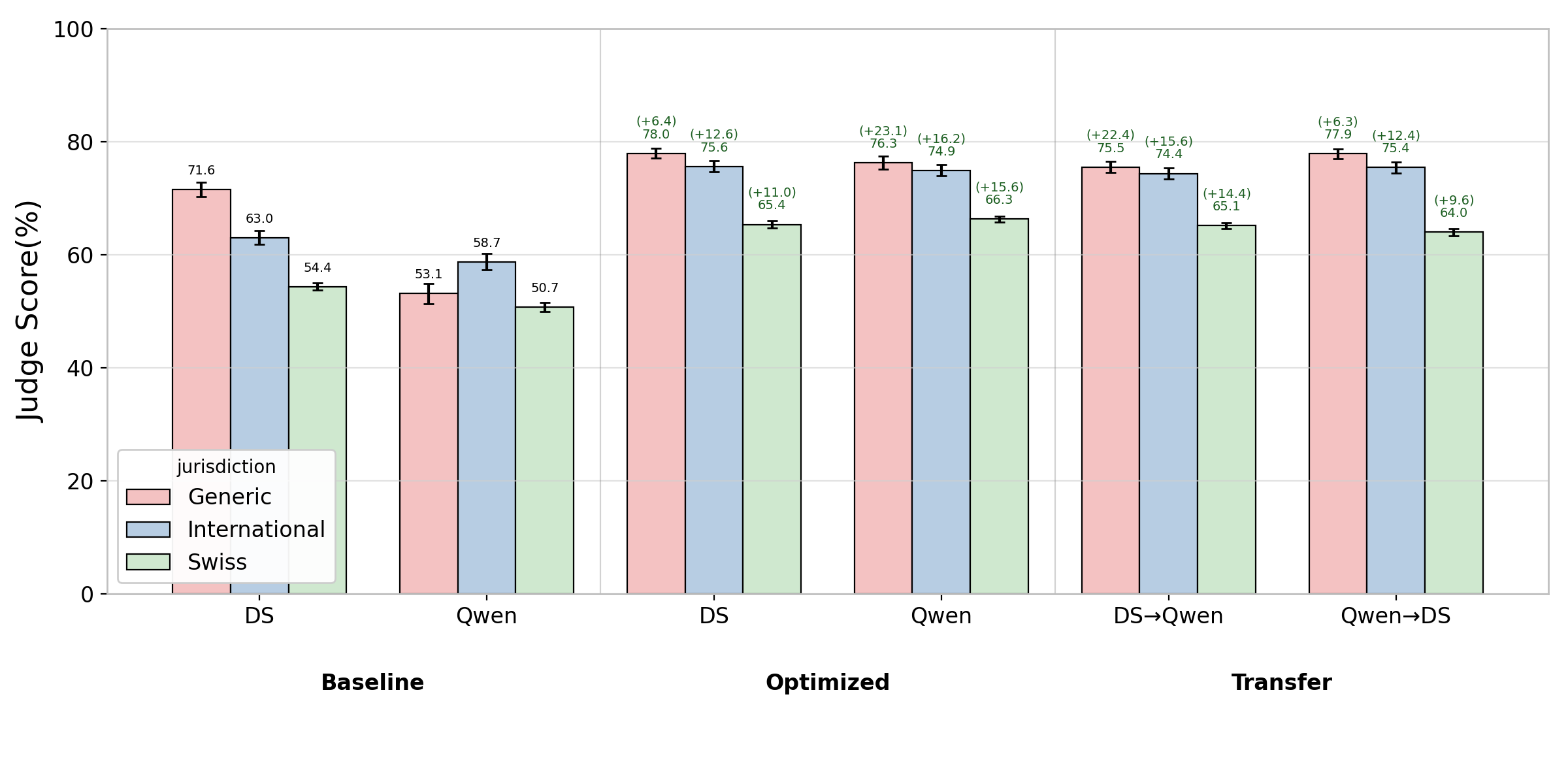}
  \caption{Jurisdiction-wise performance for qwen3-235b.}
  \Description{}
  \label{fig:jurisdiction_235b}
\end{figure}
\vspace{-0.8em}

\begin{figure}[H]
  \centering
  \captionsetup{font=normalsize, labelfont=bf, skip=2pt}
  \includegraphics[width=\linewidth,height=4.5cm,keepaspectratio]{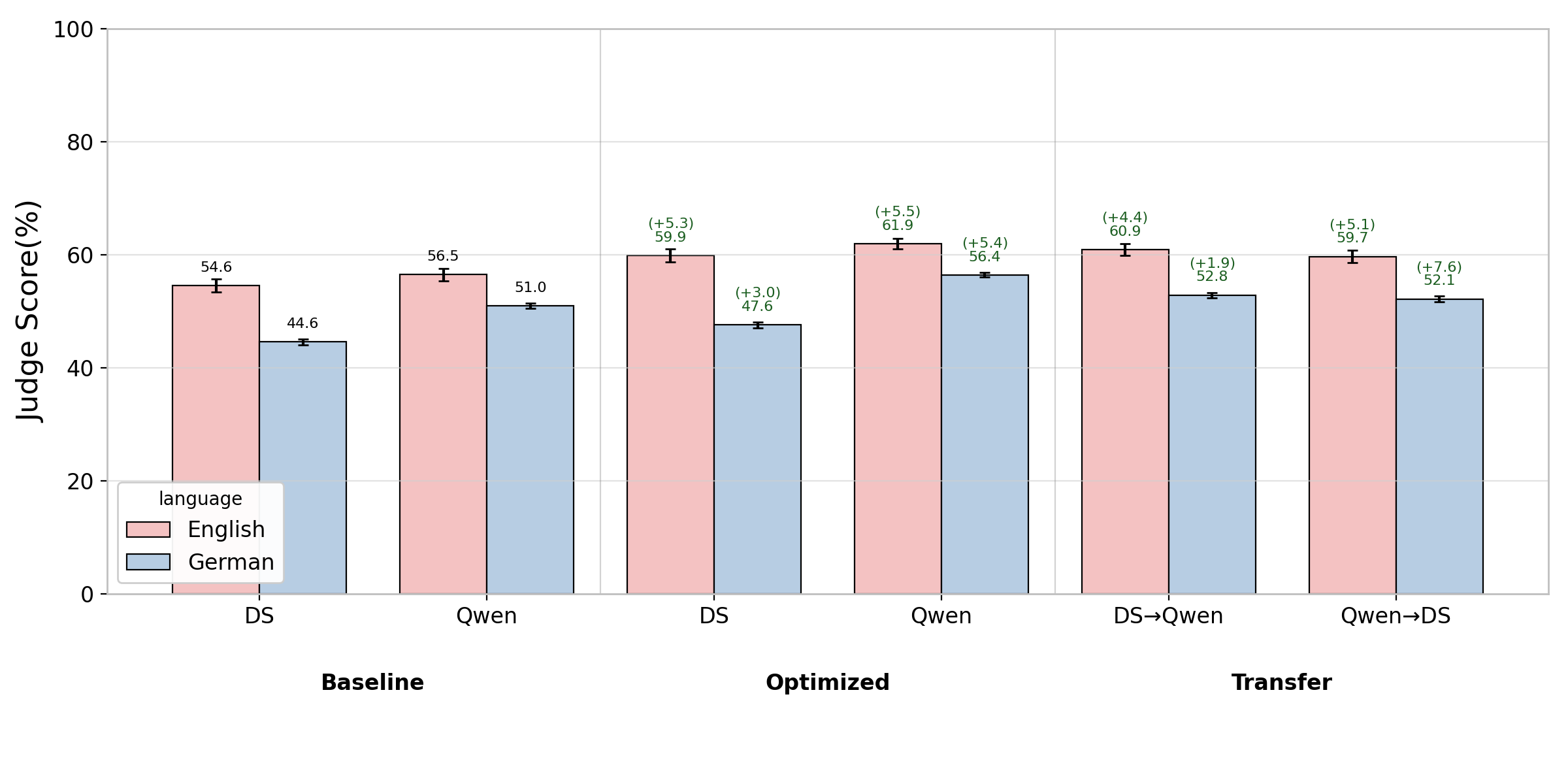}
  \caption{Language-wise performance for gpt-oss-20b.}
  \Description{}
  \label{fig:language_20b}
\end{figure}
\vspace{-0.8em}

\begin{figure}[H]
  \centering
  \captionsetup{font=normalsize, labelfont=bf, skip=2pt}
  \includegraphics[width=\linewidth,height=4.5cm,keepaspectratio]{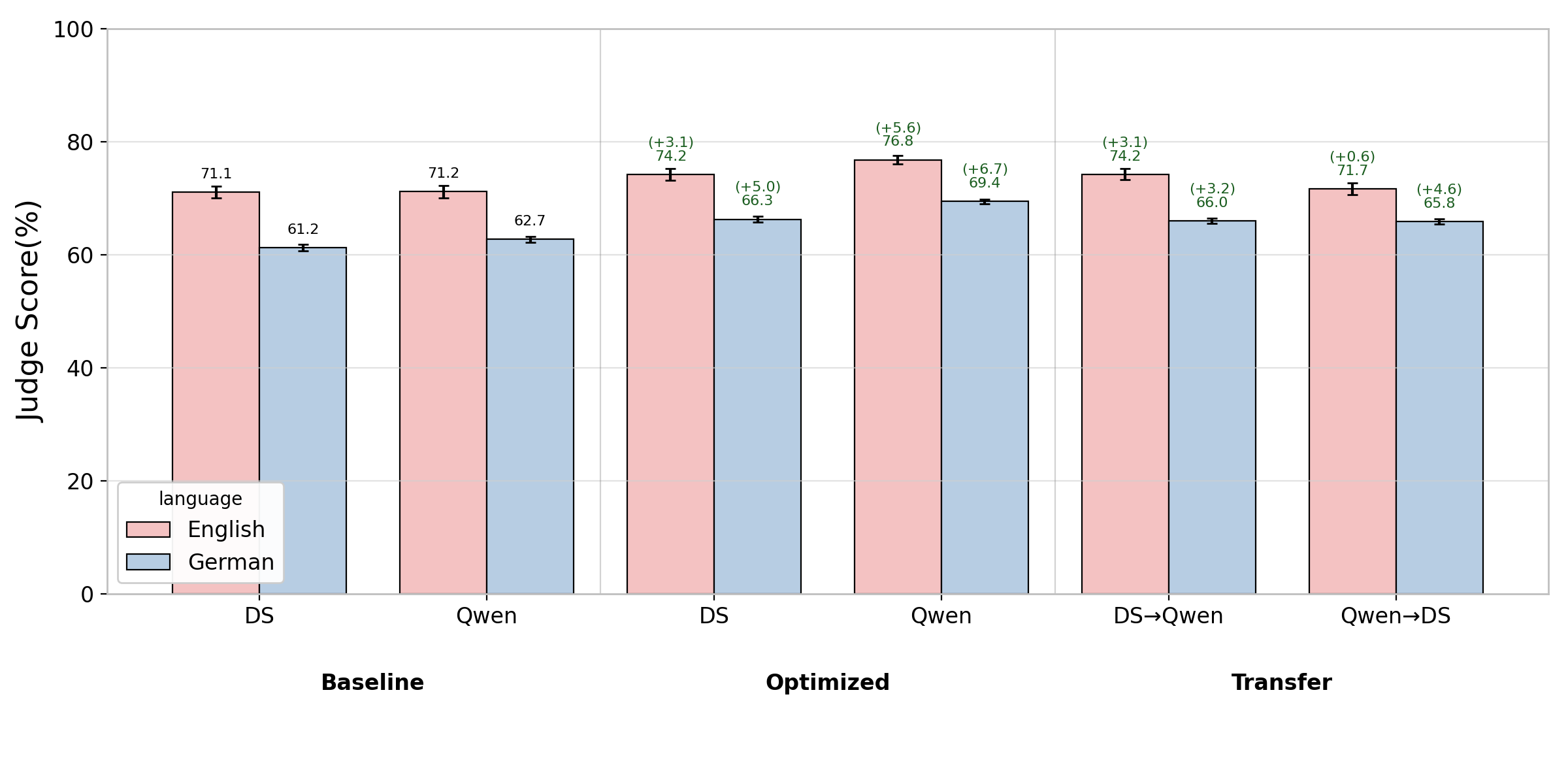}
  \caption{Language-wise performance for gpt-oss-120b.}
  \Description{}
  \label{fig:language_120b}
\end{figure}
\vspace{-0.8em}

\begin{figure}[H]
  \centering
  \captionsetup{font=normalsize, labelfont=bf, skip=2pt}
\includegraphics[width=\linewidth,height=4.5cm,keepaspectratio]{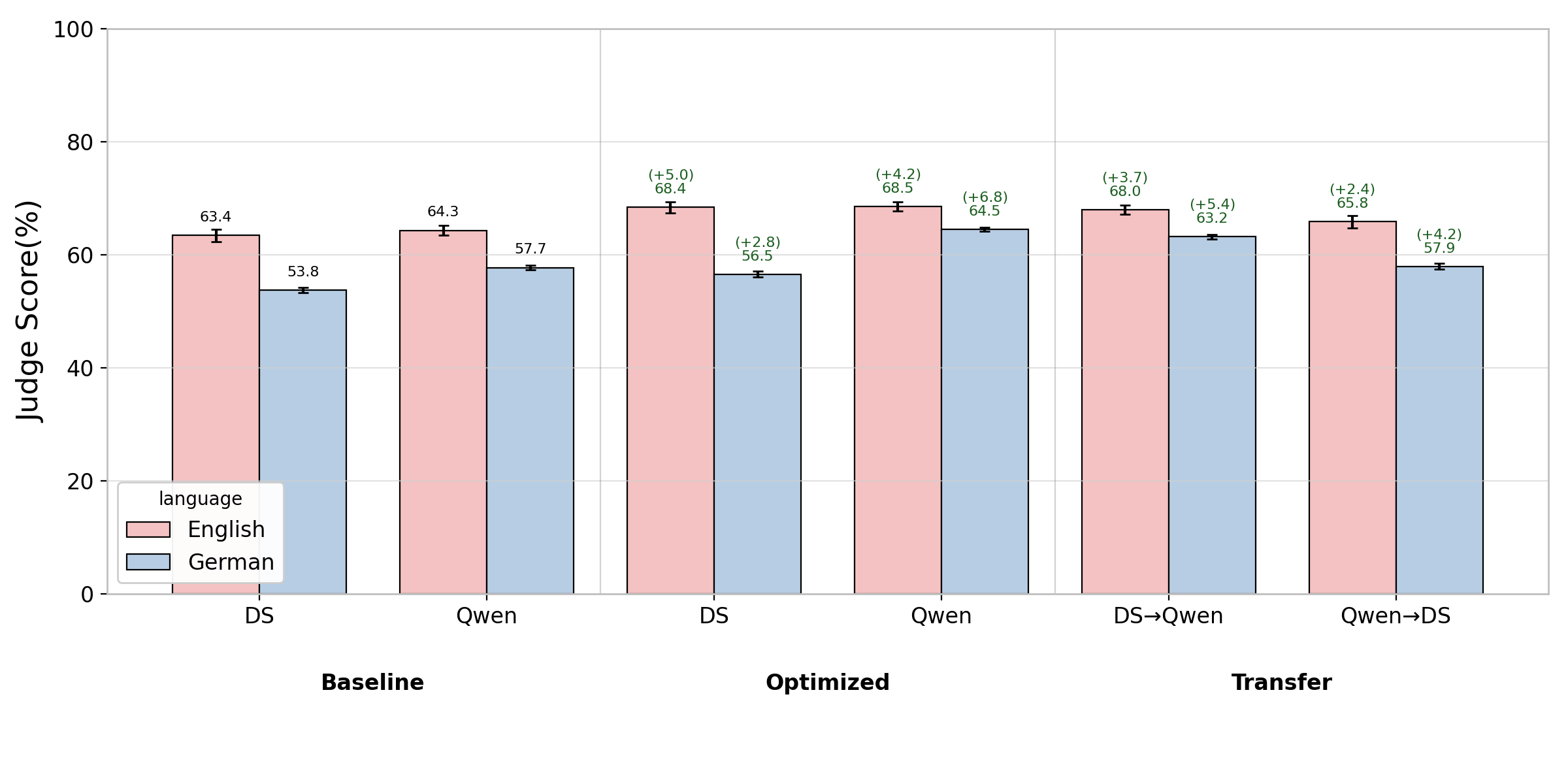}
  \caption{Language-wise performance for qwen3-32b.}
  \Description{}
  \label{fig:language_32b}
\end{figure}
\vspace{-0.8em}

\begin{figure}[H]
  \centering
  \captionsetup{font=normalsize, labelfont=bf, skip=2pt}
  \includegraphics[width=\linewidth,height=4.5cm,keepaspectratio]{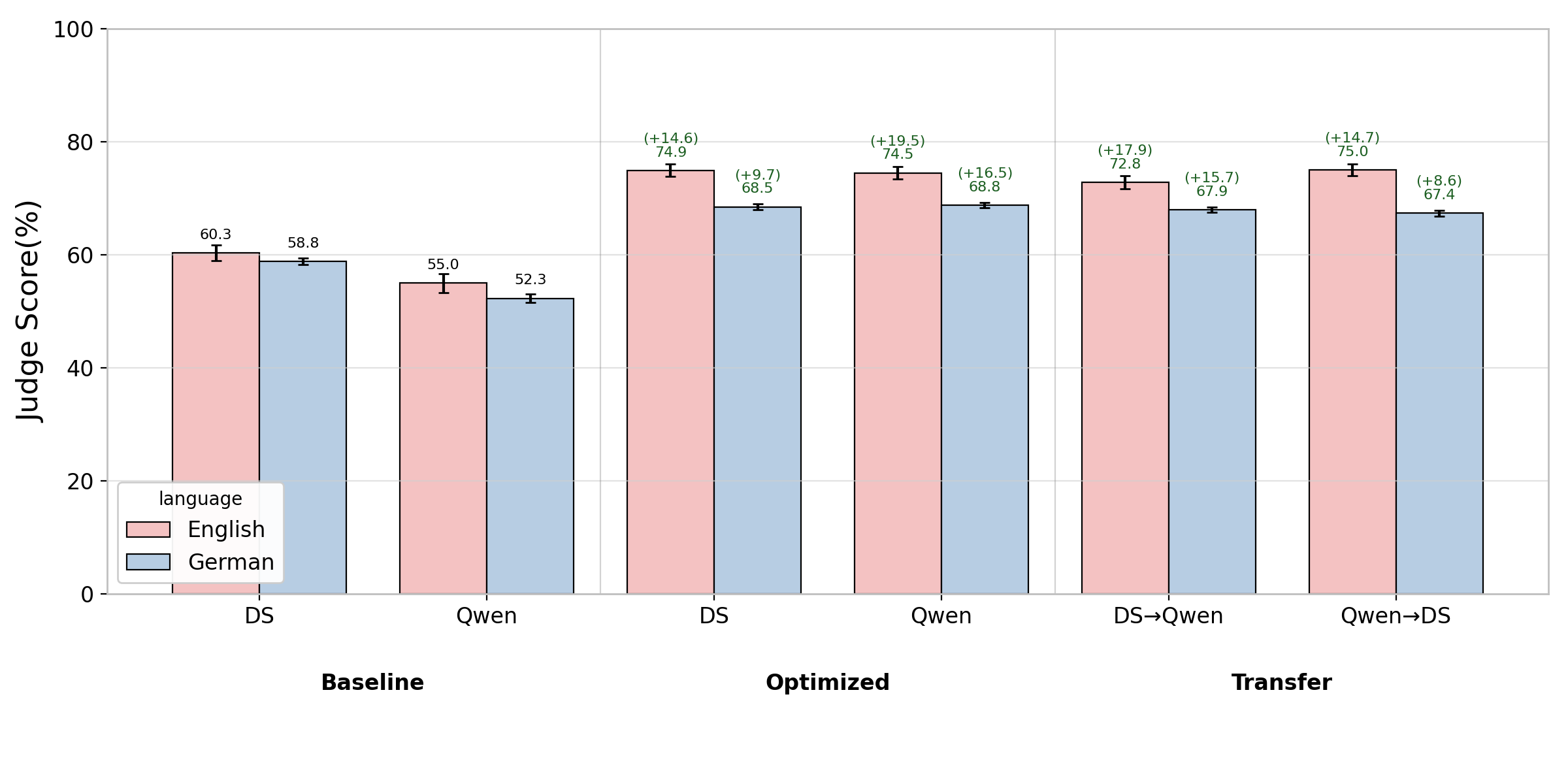}
  \caption{Language-wise performance for qwen3-235b.}
  \Description{}
  \label{fig:language_235b}
\end{figure}
\vspace{-0.8em}

\end{document}